\documentclass[a4paper,fleqn]{cas-dc}

\usepackage[authoryear,longnamesfirst]{natbib}
\usepackage{graphicx}
\usepackage{amsmath}
\usepackage{hyperref}

\newcommand{\Autoref}[1]{%
  \begingroup%
  \def\sectionautorefname{Section}%
  \def\subsectionautorefname{Subsection}%
  \def\subsubsectionautorefname{Subsubsection}%
  \def\figureautorefname{Figure}%
  \def\tableautorefname{Table}%
  \autoref{#1}%
  \endgroup%
}

\usepackage[linesnumbered,ruled,vlined]{algorithm2e}
\def\tsc#1{\csdef{#1}{\textsc{\lowercase{#1}}\xspace}}
\tsc{WGM}
\tsc{QE}
\tsc{EP}
\tsc{PMS}
\tsc{BEC}
\tsc{DE}

\begin{document}
\let\WriteBookmarks\relax
\def\floatpagepagefraction{1}
\def\textpagefraction{.001}

\shorttitle{MA-DARTS}

\shortauthors{Yilei Man et~al.}

\title [mode = title]{Differentiable architecture search with multi-dimensional attention for spiking neural networks}                      



%
\author[1,2,3]{Yilei Man}[
                        orcid=0009-0005-2665-1957]
\credit{Data collection and curation, Conceptualization, Methodology, Software, Investigation, Resources, Writing - Original draft preparation \& editing}

\ead{manyilei21@mails.ucas.ac.cn}



\affiliation[1]{organization={Nanjing Institute of Intelligent Technology},
    city={Nanjing},
    postcode={211100}, 
    country={China}}
\affiliation[2]{organization={School of Integrated Circuits, University of Chinese Academy of
Sciences},
    city={Beijing},
    postcode={100080}, 
    country={China}}
\affiliation[3]{organization={University of Chinese Academy of
Sciences (UCAS)},
    city={Beijing},
    postcode={100049}, 
    country={China}}
\affiliation[4]{organization={Institute of Microelectronics, Chinese Academy of Sciences},
    city={Beijing},
    postcode={100029}, 
    country={China}}

\author[1,2,3]{Linhai Xie}[style=chinese]
\credit{Writing - review, Investigation, Methodology}
\author[1,4]{Shushan Qiao}[style=chinese]
\credit{Writing - review, Supervision, Funding acquisition}
\author[1,4]{Yumei Zhou}[style=chinese]
\credit{Writing - review, Supervision, Funding acquisition}
\author[1,4]{Delong Shang}[%
   ]
\ead{shangdelong@ime.ac.cn}
\cormark[1]

\credit{Supervision, Funding acquisition, Project administration}

\cortext[cor1]{Corresponding author}



\begin{abstract}
\noindent Spiking Neural Networks (SNNs) have gained enormous popularity in the field of artificial intelligence due to their low power consumption. However, the majority of SNN methods directly inherit the structure of Artificial Neural Networks (ANN), usually leading to sub-optimal model performance in SNNs. To alleviate this problem, 
we integrate Neural Architecture Search (NAS) method and propose Multi-Attention Differentiable Architecture Search (MA-DARTS) to directly automate the search for the optimal network structure of SNNs. 
Initially, we defined a differentiable two-level search space and conducted experiments within micro architecture under a fixed layer. Then, we incorporated a multi-dimensional attention mechanism and implemented the MA-DARTS algorithm in this search space. Comprehensive experiments demonstrate our model achieves state-of-the-art performance on classification compared to other methods under the same parameters with 94.40\% accuracy on CIFAR10 dataset and 76.52\% accuracy on CIFAR100 dataset. Additionally, we monitored and assessed the number of spikes (NoS) in each cell during the whole experiment. Notably, the number of spikes of the whole model stabilized at approximately 110K in validation and 100k in training on datasets.
\end{abstract}




\maketitle

\section{Introduction}





Spiking Neural Networks (SNNs) have been widely explored in the artificial intelligence field in recent years due to their sparsity, low power consumption, and their ability to mimic dynamic neuronal properties, earning them the title of "third-generation neural networks" \cite{maass1997networks}.
However, the existence of binary spikes makes training SNNs a challenging task. 

Existing SNN training algorithms can be roughly divided into two categories: those based on neurobiological characteristic training algorithms, and those adapted from traditional neural network training methods for spiking neural networks. Among the former, methods represented by Spike Timing-Dependent Plasticity(STDP) \cite{caporale2008spike} have achieved remarkable results only in shallow networks for lacking global information \cite{iakymchuk2015simplified}. Therefore, most algorithms have been enhanced based on the work of traditional artificial neural networks(ANNs). However, directly inheriting network structures from ANN will inevitably lead to accuracy loss in deep spiking neural networks. Many researchers \cite{lee2020enabling}, \cite{sengupta2019going}, \cite{hu2021spiking},\cite{han2020rmp}, \cite{fang2021deep} have developed improved residual networks considering aspects like residual block design, compensation mechanisms, and new neuron models, demonstrating decent performance in terms of accuracy, parameters and time steps. However, these methods, which are artificially designed sub-optimal network structures within a fixed framework, still lag behind traditional neural networks of the same structure in terms of performance.

Neural Architecture Search (NAS) first proposed by \cite{zoph2016neural} is an automated approach for designing effective neural networks, which often achieves more remarkable results than artificially designed networks in many scenarios \cite{xie2021weight}. Therefore, this method can also be applied to exploit the potential of the network structure of SNN. In this work, we introduce a significant neural architecture search method to automate the search for the optimal network structure of SNNs within a specific search space.

Specifically, we initially established a search space consisting of two levels, including macro backbone architecture and micro candidate architecture. To effectively find the optimal network structure in this search space and consider the temporal information of SNN when processing event-driven data, we proposed a Multi-Attention Differentiable Architecture Search (MA-DARTS) algorithm, incorporating an attention mechanism \cite{bahdanau2014neural} at a crucial stage based on the original algorithm \cite{liu2018darts}. Unlike traditional types of attention mechanisms, this attention mechanism not only includes channel attention \cite{hu2018squeeze} and spatial attention but also temporal attention \cite{yao2022attention,yao2021temporal}. We conducted comprehensive experimental verification of the proposed method on two datasets, CIFAR10 and CIFAR100, evaluating and analyzing the network from accuracy, network parameters, and the number of spikes. The result demonstrates that MA-DARTS not only outperforms manually designed SNN network structures in terms of accuracy but also achieves more lightweight network parameters. Moreover, the addition of an attention mechanism can also bring accuracy gains to the model with a minimal number of parameters. The contributions can be summarized as follows:
\begin{enumerate}[\textbullet]
\item To simplify the search process, we first established a differentiable two-level search space and then searched for micro architecture within fixed layers. This method enables efficient search with limited computational resources.
\item We enhanced DARTS by incorporating multi-dimensional attention mechanisms within the defined search space and proposed MA-DARTS algorithm. Our algorithm effectively improved the model's classification accuracy with the addition of a minimal number of parameters. 
\item We conducted comprehensive experimental verification on CIFAR10 and CIFAR100. The results show that our method can identify network structures with superior accuracy and parameter metrics at the same network size compared to other methods.
\end{enumerate}

 The rest of the paper is organized as follows. ~\Autoref{Section2} introduces related work. ~\Autoref{Section3} provides a comprehensive overview of our methodology in details.~\Autoref{section4} presents the experimental results compared to other methods and demonstrate a in-depth analysis of connections in micro architecture.
Finally, the findings and the potential directions for future work are summarized in ~\autoref{section5}. 
 
\section{Related work}\label{Section2}
There has been ongoing research in the field of SNN using NAS method in recent years. Liquid State Machines (LSMs), which are a type of recurrent SNN, have the ability to emulate the structure and functionality of the human brain, making them more biologically realistic. However, only a few LSM models have demonstrated superior performance compared to traditional artificial neural networks when it comes to solving real-world classification or regression problems. \cite{zhou2020surrogate} proposed a powerful method combining LSM with evolutionary algorithm, proving to be efficient and effective in optimizing the parameters and architecture of the LSM. Furthermore, another study \cite{tian2021neural} introduced a three-step heuristic search technique to handle the enormous search space, and it demonstrates its effectiveness on the N-MNIST image dataset and FSDD voice dataset using the simulated annealing algorithm.

Actually, some researchers explored the combination of genetic algorithms in their studies. They designed their special fitness functions and achieved favorable SNN structures. GAQ-SNN(\cite{nguyen2022gaq}) focuses on reducing the memory weight overhead due to the constrained on-chip memory available in edge computing platforms. Results from simulations and hardware implementations indicate that GAQ-SNN can reduce memory storage by up to 12.5 times compared to the baseline network while maintaining an accuracy loss of only 0.6\%.

A notable characteristic of spiking neural networks is their efficiency in processing spatio-temporal information through discrete and sparse spikes. However, some works focus only on the accuracy of the algorithm, while neglecting the power consumption of the design. Auto-SNN \cite{na2022autosnn} analyzed the power consumption in terms of neural network structure, network pooling layer, and downsampling layer and combined with evolutionary algorithms successfully designed network structures with low spike firing rates.

There are also some significant NAS methods for designing neural architecture in SNN. SpikeDHS \cite{che2022differentiable} used multiple gradients to enhance Differentiable Architecture Search (DARTS) \cite{liu2018darts} and completed target detection in event-driven scenarios, and SNASNet \cite{kim2022neural} proposed a NAS algorithm that doesn't require training by incorporating Sparsity-Aware Hamming Distance. These NAS methods can effectively search for the optimal neural network structure, among which the family of differentiable NAS methods \cite{liang2019darts+,chen2019progressive,chu2020fair} are significantly faster than evolutionary algorithms and reinforcement learning \cite{zoph2016neural} methods in search efficiency. Despite their excellent performance, they occupy more resources in terms of time step and model size. Therefore we propose a lightweight method that almost minimizes the sacrifice of model performance to design the structure of spiking neural networks.
\section{Methodology}\label{Section3}
\subsection{Preliminary}\label{neuron}
There are many neuron models for SNNs, and Leaky Integrate and Fired (LIF) neuron model is the most widely used model \cite{gerstner2002spiking,hunsberger2015spiking}. Like other works \cite{na2022autosnn,che2022differentiable,wu2019direct}, we convert the continuous dynamic properties into a discrete iterative version, which can be described as:
\begin{equation}
u_i^{t,n}=\tau u_i^{t-1,n} (1-o_i^{t-1,n} )+x_i^{t,n},\label{1}
\end{equation}
\begin{equation}
o_i^{t,n}=Spike(u_i^{t,n}-V_{th} ),\label{2}
\end{equation}
\begin{equation}
x_i^{t,n}=\sum _{j=1}^{l(n-1)}w_{ij}^n*o_j^{t,n-1}, \label{3}
\end{equation}
In the above equation, superscript $t$ denotes the $t$-th time step, superscript $n$ denotes the $n$-th layer, subscript $i$ denotes the $i$-th neuron among a specific layer, $l(\,)$ denotes the number of neurons. $u_i^{t,n}$ denotes the accumulated membrane potential on the $i$-th neuron at the $t$-th time step and the $n$-th layer,  $o_i^{t,n}\in\{0,1\}$ denotes the output of the $i$-th neuron at the $t$-th time step and the $n$-th layer, where $o_i = 1$ denotes a neuron firing a spike and $o_i = 0$ denotes no neuronal activity, and $x_i^{t,n}$ denotes the total incoming neuronal activity from the presynaptic layer received by the $i$-th neuron at the $t$-th time step and the $n$-th layer. $w_{ij}^n$ denotes the synaptic weight from the $j$-th neurons in the previous layer to the $i$-th neurons in the next layer, $\tau$ denotes the membrane potential decay constant, $V_{th}$ denotes the threshold for neuron spike generation, and $spike(\,)$ is the neuron activation function, defined as $spike(x) = 1$ or 0 for $x>0 $ or $x<0$. In our setting, $\tau = 0.2$, $V_{th} = 0.5$. 

Each neuron at the $t$-th time step and the $n$-th layer corresponds to 2 important parameters $u_i^{t,n}$ and $o_i^{t,n}$, where $u_i^{t,n}$ is calculated by the membrane potential $u_i^{t-1,n}$ at the previous moment that fades with time and the total incoming neuronal activity from the presynaptic layer and $o_i^{t,n}$ is calculated by the effective membrane potential value passing through the neuron activation function.

\begin{figure}
\centerline{\includegraphics[width=\columnwidth]{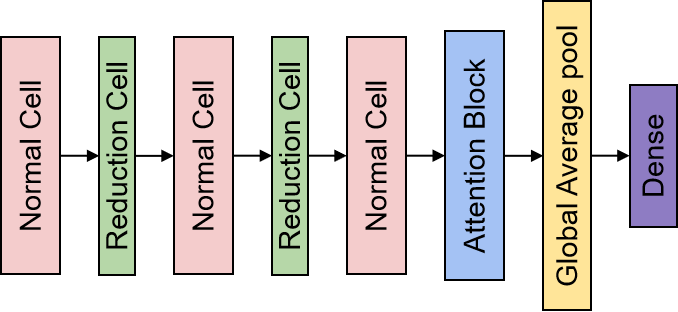}}
\caption{Macro architecture of proposed network under the DARTS search space.}\label{fig1}
\end{figure}
\subsection{The DARTS Search Space}\label{search_space}
Search space \cite{elsken2019neural} determines the scope of neural network structure, and formulating its scale is a meaningful and challenging task. A larger search space can allow the search scale to cover more neural network structures, thereby increasing the likelihood of finding more effective ones. Correspondingly, a larger search space also can make it difficult for the search algorithm to converge or require more time. Meanwhile, the scale of the search space is typically related to model evaluation, therefore determining search space is actually a complex task.

There are various types of search spaces in the NAS field, and they generally consist of two parts: macro architecture and micro architecture (Cell). Macro architecture determines the main backbone of the network while the micro architecture determines the details of the network units. Most NAS algorithms nowadays choose to fix the macro architecture and search for more remarkable micro architecture to better determine the scale of the search space. Similarly, we implemented MA-DARTS to search for the optimal spiking neural network structure in the DARTS Search Space with fixed macro architecture. The candidate operations in standard micro architecture are defined as sep-conv-3x3, sep-conv-5x5, dil-conv-3x3, dil-conv-5x5, max-pool-3x3, avg-pool-3x3, None, and skip-connect. Considering that a cell unit has 4 intermediate nodes, then both the number of possible types of reduction cell and normal cell are $C_{2}^{2} C_{3}^{2}C_{4}^{2}C_{5}^{2}\cdot7^8\approx 1.037\times10^9$. Specifically, for a network with 5 layers, the size of the entire search space is approximately $5\times10^9$. In the subsequent experiments, we specified the number of cells as 5 or 6 at the stage of search, and the corresponding macro architecture during the search process is shown in ~\autoref{fig1}.
\subsection{Multi-dimensional Attention} \label{attention}
\begin{figure}
\centerline{\includegraphics[width=\columnwidth]{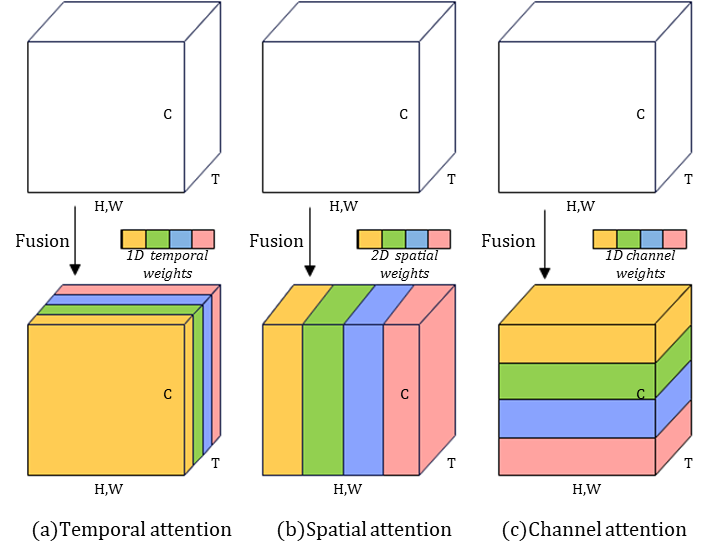}}
\caption{Different dimensions of attention.}\label{fig2}
\end{figure}
The final step of the traditional DARTS searching cell is to obtain the output by directly concatenating four nodes without distinguishing the relationships between the nodes. A previous study \cite{xu2019pc} proposed to further set a channel merging parameter similar to the structural parameter, but this approach would greatly increase the model complexity and make training more complicated. Therefore, we propose to directly insert an attention mechanism into the concatenation step, which can effectively distinguish the information of each node and only bring minimal model complexity according to our experiments. In the traditional ANN field, attention mechanisms are usually divided into channel attention, spatial attention, and channel-spatial mixed attention mechanisms. Similarly, when dealing with input that contains temporal dimension information, the type pf attention mechanisms could be extended, which leads to the emergence of temporal attention mechanisms. The principles of the three attention mechanisms are shown in \autoref{fig2}. Through different attention modules, attention weight can be obtained. Specifically, the weight of spatial attention is two-dimensional while the weight of the other two types of attention is one-dimensional. 
A general attention mechanism process can be described as follows:
\begin{equation}
X_{AT} = h(f(x),x)
\end{equation}
In this formula, $X_{AT}$ refers to a type of attention mechanism, $f(\,)$ specifies an operation function for obtaining attention mechanism weights, and $h(\,)$ denotes an operation function for applying attention mechanism weights to the original input. Since inputs often contain information of diverse dimensions, we can devise different types of attention mechanism functions to suit different input types. Drawing on this idea, we proposed MA-DARTS algorithm and designed various attention functions $f$. The design principles for two variants of the attention function $f$ are presented below.
\begin{figure}
    \centerline{\includegraphics[width=\columnwidth]{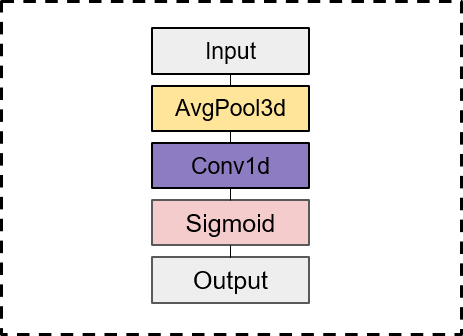}}
    \caption{The mechanism in Efficient Channel Attention (ECA) Module.}
    \label{ECA_f}
\end{figure}

\begin{enumerate}[a)]
\item{\it Channel-Temporal-DARTS(CT-DARTS)}: Based on the principle of Efficient Channel Attention \cite{wang2020eca}, we designed a module that combines channel and temporal attention mechanism as shown in \autoref{ECA_f}, which can be described as follows:
\begin{equation}
f(X^n) = \sigma(Conv_{C,T}(AvgPool_{S}(X^n)))
\end{equation}
\begin{equation}
X_{AT}^n = f(X^n) \, X^n
\end{equation}

Here, superscript $n$ denotes the $n$-th layer, $X^n =[X^{1,n}...,$\\$ X^{t,n}, ..., X^{T,n}]\in \mathbb{R}^{N\times C\times H\times W\times T}$ denotes the input feature map arranging in batch, channel, height, width and time step, and the attention function $f(\,)$ is a function that simultaneously considers both channel and temporal dimensions. The $AvgPool_{S}$ here is a pooling layer that pools over the spatial dimension, therefore left both the channel and temporal dimensions information, while the corresponding $Conv_{C,T}$ layer is a $1\times 1$ convolution layer along the channel dimension and temporal dimension with adaptive kernel sizes based on the number of channels, and $\sigma$ denotes sigmoid activation function. Finally, the original feature map $X^n$ is scaled by  $f(X^n) \in \mathbb{R}^{N\times C\times 1\times 1\times T}$ representing the importance of each channel and each time step. By this way, this module allows the model to focus more on the more important information while ignoring the less important ones.

\begin{figure}
    \centerline{\includegraphics[width=\columnwidth]{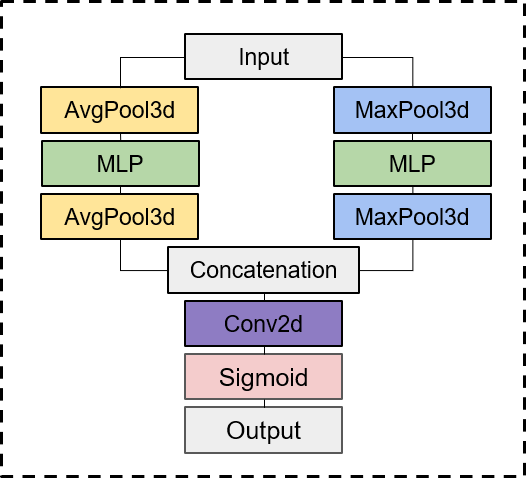}}
    \caption{The mechanism in Convolutional Block Attention (CBAM) Module.}
    \label{CBAM_f}
\end{figure}

\item{\it MA-DARTS}: Based on the principle of the Convolutional Block Attention Module \cite{woo2018cbam}, we have designed a module that sequentially obtains channel-temporal attention and spatial attention as shown in \autoref{CBAM_f}. The process of obtaining channel-temporal attention mechanism can be described as follows:
\begin{equation}\label{E1}
\begin{split}
f(X^n) = \sigma (MLP(AvgPool_{S}(X^n)) \\
 + MLP(MaxPool_{S}(X^n)))
\end{split}
\end{equation}
\begin{equation}
X_{C,T}^n = f(X^n) \cdot X^n
\end{equation}
The process of obtaining spatial attention mechanism can be described as follows:
\begin{align*}
f_S(X_{C,T}^n) = \sigma (Conv_{S}([MaxPool_{C}(X_{C,T}^n),\\
AvgPool_{C}(X_{C,T}^n)]))\tag*{(9)}
\end{align*}
\begin{equation} 
X_{AT}^n = f_S(X_{C,T}^n) \cdot X_{C,T}^n \tag*{(10)}
\end{equation}
Here, $f(X^n)$ denotes the channel and temporal attention function, and $f_S(X_{C,T}^n)$ denotes the spatial attention function. The original feature map is fed into $AvgPool_{S}$ and $MaxPool_{S}$ operation respectively. Then, the output will be sent into a MuLti-Layer Perceptron ($MLP$), where $MLP$ consists of $ReLU$ layer and 2 convolutional layers. The two results are added together and then pass through the sigmoid function to get $f(X^n)$. Both $AvgPool_{C}$ and $MaxPool_{C}$ are pooling layer that pools over channel dimension. $X_{C,T}^n$, which combines temporal attention and channel attention information, will be concatenated in channel dimension after undergoing two pooling operations. The $Conv_{S}$ is a $2\times 1$ convolution layer and the above result will pass it at each time step. At last, the importance of space represented by $f(X_{C,T}^n) \in \mathbb{R}^{N\times 1\times H\times W\times T}$ is used to scale $X_{C,T}^n$.
\end{enumerate}

\subsection{Multi-Attention Differentiable Architecture Search}
The family of differentiable NAS methods has demonstrated significant results in classification and event-based deep stereo tasks in the field of SNN. Nevertheless, most methods neglect the consideration of model size and limited resources. Inspired by this idea, we propose a lightweight NAS method aiming to show exceptional results in SNN while maintaining a small model size.

\begin{figure*}
\centerline{\includegraphics[width=2\columnwidth]{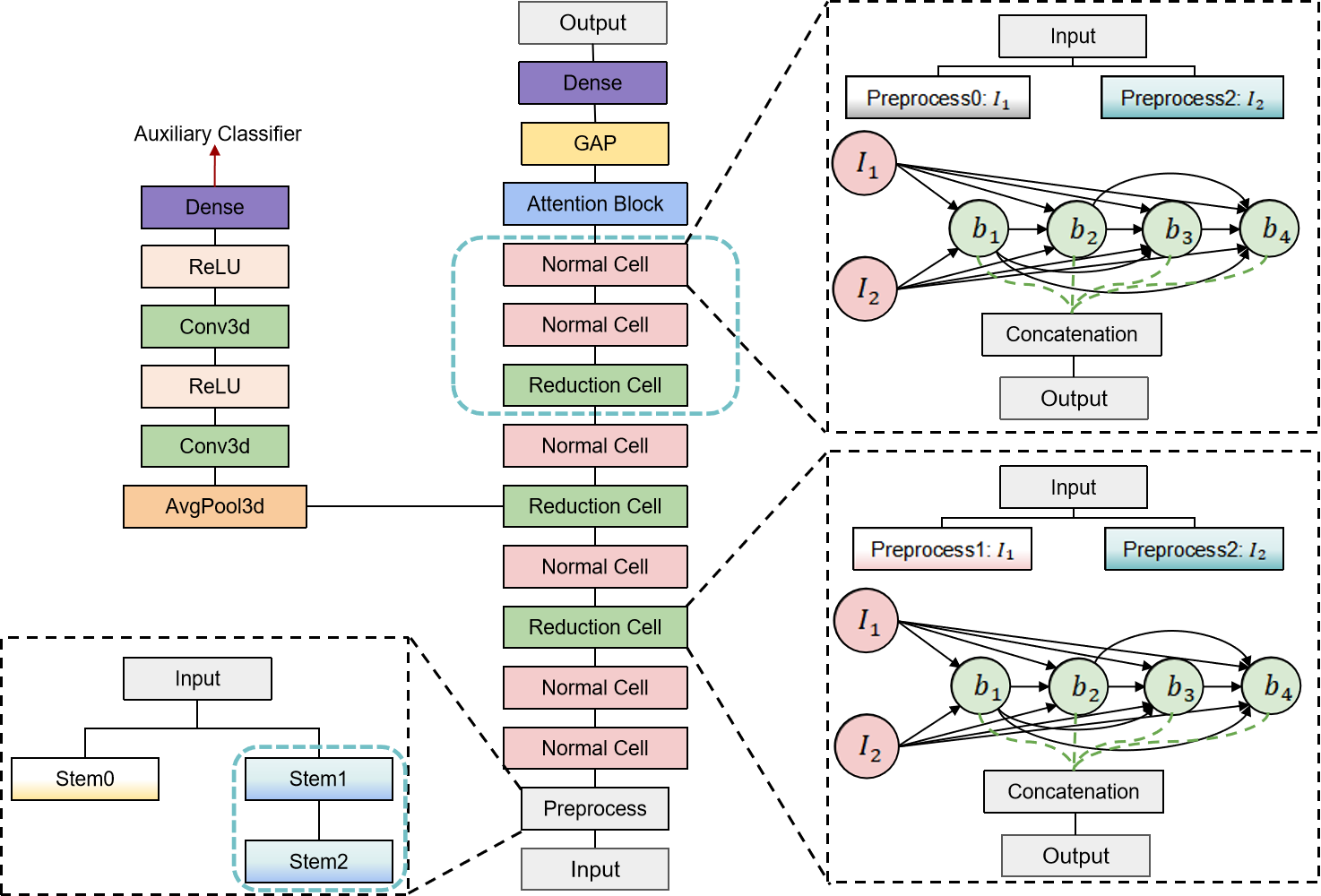}}
\caption{The overview of Multi-Attention Differentiable Architecture Search(MA-DARTS) .}\label{MA-DARTS} 
\end{figure*} 

The overview of MA-DARTS is shown as in \autoref{MA-DARTS}. The macro architecture comprises 6 layers of arranged cells and the extra layers in the blue box are prepared for higher resolution datasets. Both types of cells share similarities, with the main distinction in the preprocessing step: the reduction cell decreases the feature map size, while the normal cell preserves it. Each type of cell consists of 4 internal nodes, where the later nodes rely on the previous nodes. The connection of internal nodes can be described as below:
\begin{equation} 
b_i = \sum_{\substack{i < j}}\overline{o}^{(i,j)}{(b_j)} \tag*{(11)}
\end{equation}
\begin{equation} 
\overline{o}^{(i,j)}{(b_j)} = \sum_{\substack{o \in \mathcal{ O }}}\frac{e^{\alpha_o^{(i,j)}}}{\sum_{\substack{o'\in \mathcal{ O }}}e^{\alpha_{o'}^{(i,j)}}} o{(b_j)} \tag*{(12)}
\end{equation}
Here, $b_i$ denotes the $i$-th internal node, the notation $\overline{o}^{(i,j)}{(b_j)}$ indicates the mixed operation of the directed edge from $i$ to $j$ applied to $b_j$. The output of this operation applied to $b_j$ is denoted as $o{(b_j)}$, where the operation is chosen from the candidate set $\mathcal{ O }$. And $\alpha$ denotes the architecture parameter that can be learned on the validation dataset. To enhance gradient propagation, mitigate issues such as gradient vanishing or exploding, and expedite model convergence, we incorporate an auxiliary classifier \cite{nekrasov2019fast} after the 5th cell. This auxiliary classifier aids in better gradient flow throughout the network.
 
The pseudo code of the algorithm is summarized in Algorithm \ref{algo_madarts}. In the beginning, we initialize the neuron threshold $V_{th}$ and membrane potential decay constant $\tau$ as we describe in ~\autoref{neuron}. We then specify the size of the search space named the DARTS search space which is the macro structure of the network and the time window used for processing temporal information. Meanwhile, we stipulate our candidate set as we describe in ~\autoref{search_space}. In the process of forwarding the input, the dimensions of the data are not properly distinguished in the step of concatenation. Therefore, we directly incorporate an attention mechanism into the concatenation step and use attention mechanism functions as we describe in ~\autoref{attention}. Once we calculate corresponding loss, such as cross-entropy (CE) loss, we then update architecture parameter $\alpha$ and weight $w$ respectively based on the original DARTS algorithm. We also evaluate the model on validation dataset and the optimal network structure is obtained according to accuracy. The connections among the nodes still are mixed operations in the search stage, but the final structure comes from the internal nodes connection path of the cell with the two maximum architecture parameter $\alpha$.
After we derive the architecture based on the corresponding $\alpha$, we then need to retrain the final architecture with specified epochs. 

 \IncMargin{1em}
\begin{algorithm} \SetKwData{Left}{left} \SetKwFunction{Forward}{Forward} \SetKwFunction{Attention}{Attention} \SetKwFunction{Task}{Task}\SetKwInOut{Input}{input}\SetKwInOut{Output}{output}\label{algo}
	
	\Input{Training set, training epoch $E$, training iteration $I$, neuron threshold $V_{th}$ and membrane potential decay constant $\tau$}  
	 \BlankLine 
	 
       {Initialization:Search space $\mathcal{ S }$, candidate operation set $\mathcal{ O }$} and time window $T$\;  
	 \For{$e\leftarrow 1$ \KwTo $E$}{ 
            {Processing data $D$ and labels}\;
	 	\For{$i\leftarrow 1$ \KwTo $I$}{\label{forins}            
            $X$ $\leftarrow$ \Forward{$D$}\; 
	 	$X_{AT}$ $\leftarrow$ \Attention($X$)\; 
            Calculate loss on \Task($X_{AT}$)\;
	 	{Update architecture parameter $\alpha$ by descending Loss function on validation dataset}\; 
            {Update weight $w$ by descending Loss function on training dataset}\;
            {Evaluate model on validation dataset}\;
 	 	}
       }
        Derive the final architecture based on the learned $\alpha$.\;
        //Retrain the final architecture\;
        \For{$e\leftarrow 1$ \KwTo $E$}{
            {Processing data $D$ and labels}\;
            \For{$i\leftarrow 1$ \KwTo $I$}{
                // forward data on fixed architecture\;
                $X$ $\leftarrow$ \Forward{$D$}\; 
	 	  $X_{AT}$ $\leftarrow$ \Attention($X$)\;
                Calculate loss on \Task($X_{AT}$) \;
                {Update weight $w$ by descending Loss function on training dataset}\;
                {Evaluate model on validation dataset}\;
                }
        }
 	 \caption{Multi-Attention Differentiable Architecture Search (MA-DARTS)}
 	\label{algo_madarts} 
 	\end{algorithm}
 \DecMargin{1em}
 
\section{Experimental results and analysis}\label{section4}
\begin{table*}[]
\begin{center}
\caption{COMPARISON WITH STATE-OF-THE-ART MODELS ON CIFAR10}
\label{table1}
\setlength{\tabcolsep}{3pt}
\begin{tabular}{c||c|c|c|c|c}
NetWork                      & Method & Accuracy(\%)  & Channel & T & Params(M) \\
 \hline\hline
CIFARNet-\cite{wu2019direct}   & STDP & 84.36    & 16      & 8         & 0.71 \\
CIFARNet-Wu   & STDP & 86.62    & 32      & 8         & 2.83 \\
CIFARNet-Wu   & STDP & 87.80    & 64      & 8         & 11.28 \\
ResNet19-\cite{zheng2021going}  & STDP-tdBN & 89.51    & 32     & 6         & 0.93 \\
ResNet19-Zheng  & STDP-tdBN & 90.95    & 64     & 6         & 3.68 \\
ResNet19-Zheng  & STDP-tdBN & 93.07    & 128     & 6         & 14.69 \\
ResNet11-\cite{lee2020enabling}     & spike-based BP & 84.43    & 16      & 50        & 1.17 \\
ResNet11-Lee     & spike-based BP & 87.95    & 32      & 50        & 4.60 \\
ResNet11-Lee     & spike-based BP & 90.24    & 64      & 50        & 18.30 \\
CIFARNet-\cite{fang2021incorporating} & spike-based BP & 86.05    & 32     & 8         & 0.60  \\
CIFARNet-Fang & spike-based BP & 90.83    & 64     & 8         & 2.34  \\
CIFARNet-Fang & spike-based BP & 92.33    & 128     & 8         & 9.23  \\
ResNet19-\cite{lian2023learnable}       & spike-based BP  & 95.17    & -       & 2      & -      \\
VGG19-\cite{chen2022adaptive}      & conversion & 93.97    & -       & 400       & -      \\
Resnet20-\cite{kim2018deep}    & conversion & 91.40    & -       & -         & -      \\
ResNet20-\cite{han2020rmp}     & conversion  & 91.36    & -       & 2048      & -      \\
VGG16-\cite{han2020rmp}        & conversion  & 93.63    & -       & 2048      & -      \\
ResNet18-\cite{hu2023fast}       & conversion  & 95.57    & -       & 7      & -      \\
VGG19-\cite{bu2023optimal}       & conversion  & 94.95    & -       & 8      & -      \\
VGG16-\cite{kundu2021spike}    & Hybrid& 91.69    &         & 100       &        \\
SpikeDHS-\cite{che2022differentiable}    & NAS & 94.68    & 256     & 6         & 14    \\
AutoSNN-\cite{na2022autosnn}         & NAS & 91.32    & 32      & 8         & 1.46  \\
AutoSNN                 & NAS & 92.54    & 64      & 8         & 5.44  \\
AutoSNN                & NAS & 93.15    & 128     & 8         & 20.92 \\
SNASNet-FW-\cite{kim2022neural}  & NAS & 93.64    & 256     & 8         & -      \\
SNASNet-BW-\cite{kim2022neural}  & NAS & 94.12    & 256     & 8         & -      \\
CT-DARTS                 & NAS & 91.88    & 32      & 2         & 1.069 \\
CT-DARTS                  & NAS & 94.30    & 64      & 2         & 2.791 \\
CT-DARTS                 & NAS & 94.21    & 128     & 2         & 9.371 \\
CT-DARTS                 & NAS & 95.21    & 64      & 6         & 2.915 \\
MA-DARTS                      & NAS & 92.05    & 32      & 2         & 1.128 \\
MA-DARTS                      & NAS & 94.40    & 64      & 2         & 3.011 \\
\hline
\end{tabular}
\end{center}
\end{table*}

\begin{table*}[]
\caption{COMPARISON WITH STATE-OF-THE-ART MODELS ON CIFAR100}
\label{table2}
\setlength{\tabcolsep}{3pt}
\begin{tabular}{{c||c|c|c|c|c}}
NetWork        & Method & Accuracy(\%)  & Channel & T & Params(M) \\
 \hline\hline
ResNet19-\cite{lian2023learnable}       & spike-based BP  & 76.32    & -       & 2      & -      \\
ResNet20-\cite{han2020rmp} & conversion& 67.82    & -       & 2048      & -      \\
VGG16-\cite{han2020rmp}    & conversion& 70.93    & -       & 2048      & -      \\
VGG19-\cite{chen2022adaptive}   & conversion& 73.58    & -       & 1100      & -      \\
Resnet32\cite{kim2018deep} & conversion& 68.56    & -       & -         & -      \\
Resnet44-\cite{hu2021spiking}  & conversion& 70.80     & -       & -         & -      \\
SNASNet-BW-\cite{kim2022neural}  & NAS & 73.04    & 128     & 5         & 20.62 \\
SNASNet-FW-\cite{kim2022neural}  & NAS  & 70.06    & 128     & 5         & 20.54 \\
SpikeDHS-\cite{che2022differentiable}] & NAS & 76.03    & 256     & 6         & 12    \\
CT-DARTS       & NAS & 75.82    & 64      & 2         & 3.106 \\
MA-DARTS           & NAS & 76.52    & 64      & 2         & 3.231 \\
\hline
\end{tabular}
\end{table*}
\subsection{Dataset}
The data in CIFAR10 and CIFAR100 datasets consists of 50k/10k images respectively with a resolution of 32×32. A standardized normalization preprocessing step is applied to all datasets. In the experiments, each dataset is divided into three parts for training, testing, and validating. During the first stage, we maintain a 1:1 ratio between the training and testing sets. In the second stage, we adjust the ratio between the training and validation sets to 5:1.
\subsection{Implementation details}
The training process of our experiment consists of two steps. In the search stage, we adopt 5 layers of cells, each containing 4 intermediate nodes. We employ the same candidate operation set as described before and uniformly use LIF $Spike(\,)$ as the activation function. The search stage lasts for 40 epochs with a batch size of 64. Regarding weight optimization, we utilize SGD optimizer with momentum of 0.9 and cosine annealing learning rate of 0.0050. As for optimizing the architecture parameter $\alpha$, we use adam optimizer with learning rate of 0.0003. With 16 initial channels and 2 time windows, the search process takes approximately 2 GPU days on 2 NVIDIA 3090 (24GB) GPUs.

After completing the search, refinement is required. We train for 600 epochs with initial channel expansion and different batch sizes. We use SGD optimizer with momentum of 0.9 and Cosine Annealing learning rate of 0.0025. The number of cells we set varies between 5 or 6 depending on the dataset. In this process, we deploy different attention mechanisms behind the final cell to obtain the weight of different nodes. With 64 initial channels and 2 time windows, the refinement process takes 2.08 GPU days on 2 NVIDIA 3090 (24GB) GPUs.
\begin{figure*}
\centerline{\includegraphics[width=2\columnwidth]{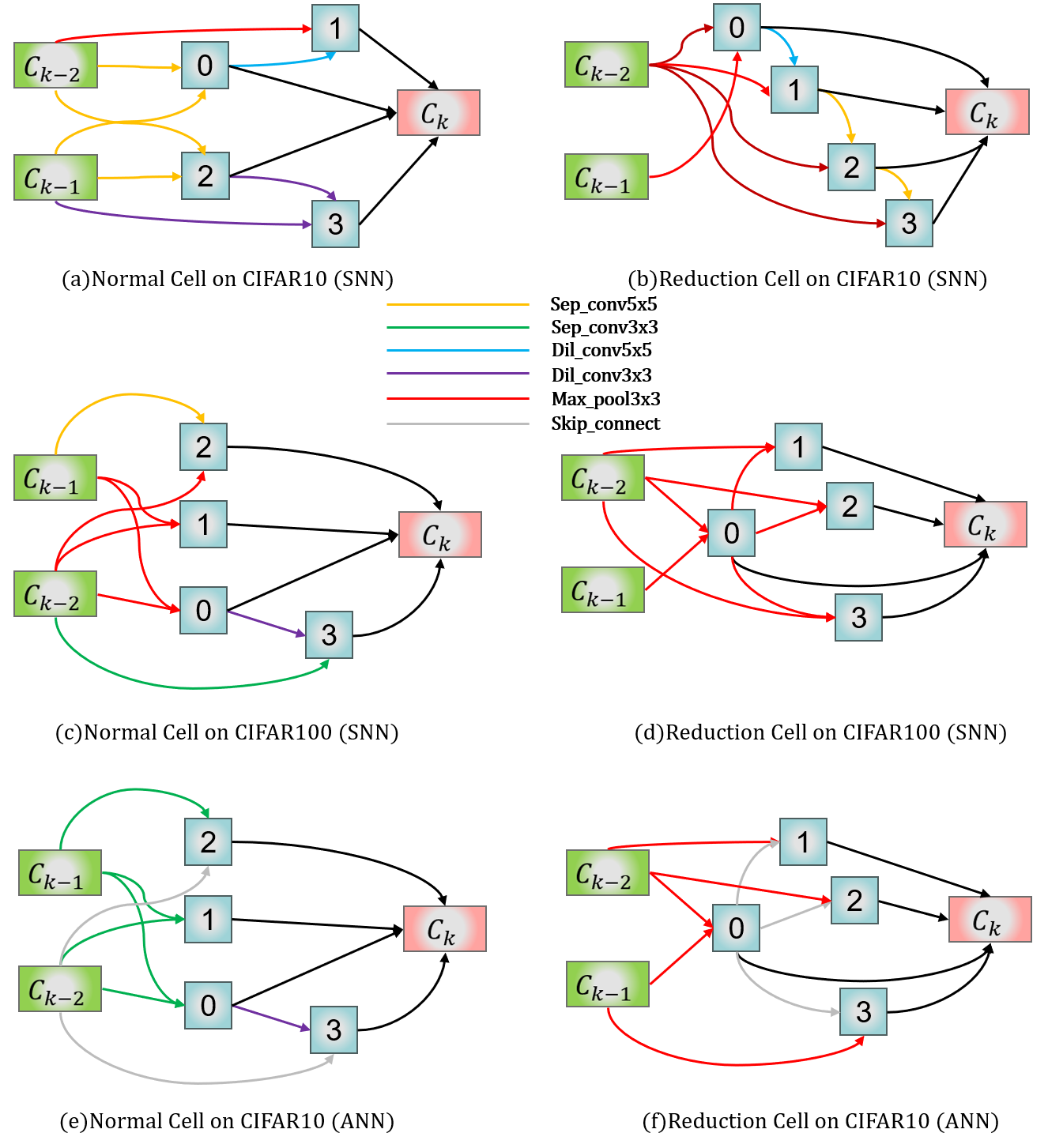}}
\caption{Final architecture for classification on CIFAR10 and CIFAR100.}\label{fig3}
\end{figure*}

\subsection{Results}
Works \cite{lee2020enabling,wu2019direct,fang2021incorporating,zheng2021going} represent a type of direct training SNN method using spike-based approaches, while studies \cite{han2020rmp,chen2022adaptive,kim2018deep} represent a type of ANN-to-SNN conversion method. ~\autoref{table1} and ~\autoref{table2} present a comparison of different methods in terms of accuracy, time steps, and parameters on the CIFAR10 and CIFAR100 datasets with various initial channels. In CIFAR10, we observe that both spike-based methods and conversion methods are closely related to the time steps and initial channels, but their performance is not satisfactory in terms of accuracy. Compared with NAS methods, our proposed method yields superior performance under identical parameter settings. Specifically, our model achieves an accuracy of 94.21\% with CT-DARTS and 94.40\% with MA-DARTS in CIFAR10 and 75.82\% with CT-DARTS and 76.52\% with MA-DARTS in CIFAR100 under 64 initial channels and 2 timesteps. Additionally, our model not only provides high classification accuracy but also has a significantly smaller parameter size compared to other methods. Among work until 2023, although SpikeDHS presents state-of-the-art accuracy performance in CIFAR10, our method can achieve similar results by extending the time steps. Specifically, our method exceeds other methods and attains an accuracy of 95.21\% on the CIFAR10 dataset with 64 initial channels and 6 time steps. Compared to work in 2023 \cite{hu2023fast,lian2023learnable,bu2023optimal}, the performance gap between these methods is not obvious, but inheriting directly from the deep classical network architecture will result in a large number of parameters. In a word, as the initial channels and time steps grow, our model's accuracy can be enhanced while maintaining a lighter model parameter size compared to other methods. 

We provide the final architecture of normal cell and reduction cell our method automatically searches in the first stage on CIFAR10 and CIFAR100 as shown in ~\autoref{fig3} (a)$\sim$(d). For both kinds of cells, there is no jump connection and average pooling operation among internal nodes. And most of the connections among internal nodes in the reduction cell are max pooling operation other than average pooling operation. Actually, this corresponds to the fact that max pooling operation can be better to preserve the asynchronous characteristic of neuron firing \cite{fang2021incorporating}. In previous studies, the types of down sampling layer are convolutional block \cite{zheng2021going}, average pooling \cite{wu2019direct} and max pooling \cite{fang2021incorporating}. However, these down sampling layers have different influences on the number of spikes and model accuracy, where average pooling operation is so hard to generate a spike that it is discouraged for the purpose of down sampling, but max pooling operation is energy-efficient in transmitting spike while maintaining a remarkable model accuracy similar to convolutional block \cite{na2022autosnn}.

And we also provide the architecture of two types of cell our method search on the ANN search space as shown in ~\autoref{fig3} (e)$\sim$(f). We first use our proposed method to search the optimal ANN network, then we use this optimal architecture to directly train on CIFAR10. And the final accuracy on CIFAR10 is only 91.32$\%$, which lags behind the majority of existing results. There are many distinctions between the ANN network structure and SNN network structure, but the most obvious distinction is that there are skip connections in the ANN network structure, which may ensure the improvement on the deep network. How to add skip connection to the network structure of SNN is a promising direction worth studying.

\begin{figure*}
\centerline{\includegraphics[width=2\columnwidth]{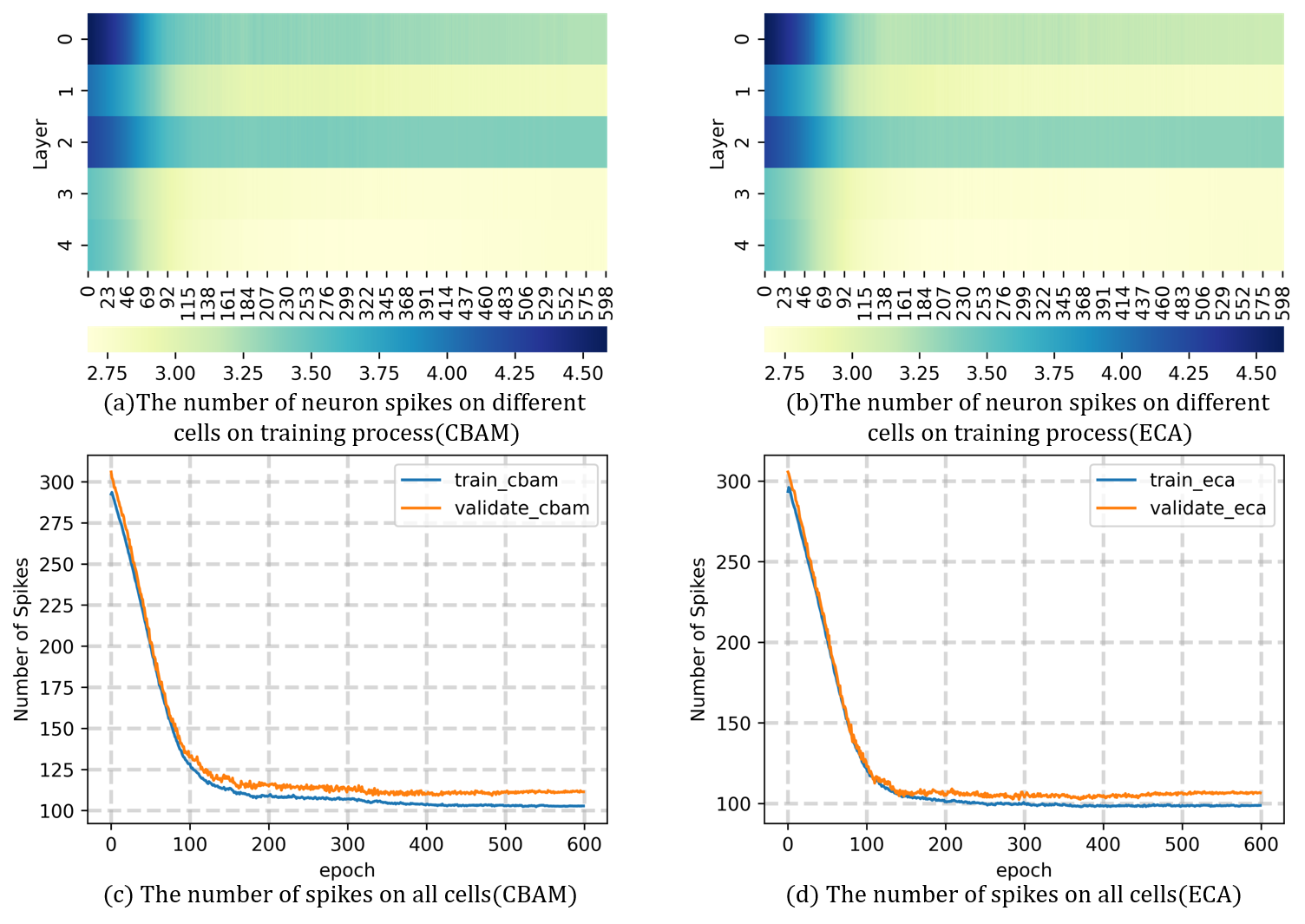}}
\caption{The number of neuron spikes on different cells and the number of spikes on all cells.}\label{fig4}
\end{figure*}

\subsection{Ablation study}
~\autoref{table3} show the results of our ablation study, where we validated and compared the effects of two attention mechanism functions on CIFAR10 and CIFAR100 datasets. The search space and experiment settings of the search process in all experiments were the same. Compared to the common DARTS algorithm, we find that both attention functions can further improve the performance of the original DARTS algorithm, and the CBAM-based attention function has a more remarkable effect. Specifically, the ECA-based attention function has a 0.51\% improvement and a 0.82\% improvement under 64 initial channels on two datasets respectively. And the CBAM-based attention function has a 0.61\% improvement and a 1.52\% improvement under 64 initial channels on two datasets respectively. Additionally, our proposed method produces a negligible increase in terms of the network's parameter size. Notably, under 2 time steps, the ECA-based model's parameter size with different initial channels barely changes but comparable classification results are obtained.

We also validated our proposed model on a more complicated dataset, like TinyImageNet-200 which has higher resolution and more categories. The micro architecture was migrated from the CIFAR10 dataset and the number of internal nodes was set to 3. The results show that MA-DARTS can achieve 59.39\% accuracy while the original DARTS algorithm lags behind this result, demonstrating the design of attention block can alleviate the difficulty of algorithm convergence.

\begin{table}[h]
\caption{COMPARISON WITH ORIGINAL DARTS}
\label{table3}
\setlength{\tabcolsep}{3pt}
\begin{tabular}{c||c|c|c|c|c}
Network      & Accuracy(\%)  & Ch & T & Params & Dataset \\
 \hline\hline
DARTS        & 92.17    & 36      & 2         & 1.227M  & CIFAR10 \\
DARTS        & 93.79     & 64      & 2         & 2.791M & CIFAR10\\
CT-DARTS & 92.90     & 36      & 2         & 1.227M & CIFAR10 \\
CT-DARTS & 94.30     & 64      & 2         & 2.791M & CIFAR10\\
MA-DARTS     & 93.03     & 36      & 2         & 1.301M & CIFAR10\\
MA-DARTS     & 94.40     & 64      & 2         & 3.011M & CIFAR10\\
DARTS        & 75.00    & 64      & 2         & 3.106M & CIFAR100\\
CT-DARTS & 75.82     & 64      & 2         & 3.106M & CIFAR100\\
MA-DARTS     & 76.52     & 64      & 2         & 3.231M & CIFAR100\\
DARTS        & 39.49    & 48      & 2         & 5.088M & TinyImageNet\\
CT-DARTS & 41.28     & 48      & 2         & 5.088M & TinyImageNet\\
MA-DARTS     & 59.39     & 48      & 2         & 5.159M & TinyImageNet\\
\hline
\end{tabular}
\end{table}



\subsection{Neural architecture analysis}
One of the salient features of spiking neural networks is low power consumption \cite{roy2019towards}. The measurement and comparison of power consumption can be done by calculating the number of neuron spikes \cite{taherkhani2020review}. In CIFAR10 dataset, we observed the changes in the number of neuron spikes (logarithmic) across different cells as the experiment progressed. ~\autoref{fig4} (a) and (b) record the number of spikes on each cell with two attention functions, which reveals that the number of spikes in all cells decreases rapidly in the first 100 epochs and stabilizes after 200 epochs. Additionally, compared to the normal cell, the reduction cell yields significantly more spikes, suggesting it consumes more power. However, after the data path passes the reduction cell, the number of spikes on normal cells will continue to decrease. This may prevent the total number of spiking from increasing. Meanwhile, we computed the number of spikes for all cells, as shown in ~\autoref{fig4} (c) and (d), and the final model stabilized at about 110K spikes on validation and 100k spikes on training.


\section{Conclusion and future works}\label{section5}
In this article, we initially introduced a differentiable two-level search space and defined the corresponding search operation set. Then we conducted the searching process within the cell under a fixed layer level and implemented the MA-DARTS algorithm in the DARTS Search Space. This method not only yields more efficient and lightweight network structures than those designed artificially but also results in outperforming other NAS algorithms under similar conditions by utilizing multidimensional attention mechanisms without significantly increasing the size of parameters. Nonetheless, our approach also needs some improvement. For example, power-aware methods can be integrated into the search process and loss function to design more universally low-power neural network structures. Based on our network architecture analysis, we also find that our method has a slight gap between validation and training, which could be solved by modifying the connection relationship of nodes and redefining the set of operations. In addition, our method has only been studied in tasks such as classification, and further research is needed in event-driven task scenarios.

\printcredits

\section*{Acknowledgements}
This work was supported by Science and Technology Innovation 2030 Project of China(2021ZD0200300). 
\bibliographystyle{cas-model2-names}

\bibliography{revision}




\bio{author3}
\textbf{Delong Shang} received the B.S. degree in computer architecture from Nanjing University, the M.S. degree in computer architecture from the Chinese Academy of Sciences, China, and the Ph.D. degree in microelectronics and computer architecture from Newcastle University, U.K. He is currently a Full Professor leading the Nanjing Institute of Intelligent Technology, China. His research interests include computer architecture, asynchronous systems, power-efficient systems, and neuromorphic computing. 
\endbio
\bio{author1}
\textbf{Yilei Man} received the B.Eng. degree in electronic engineering department from Tsinghua University, Beijing, China, in 2021. He is now a master in School of Integrated Circuits, University of Chinese Academy of Sciences. His current research interests include spiking neural network, neuromorphic computation, and artificial intelligence. 
\\
\endbio
\bio{author2}
\textbf{Lihai Xie} received the B.Eng. degree from Hangzhou Dianzi University, School of Computer Science Beijing in 2020. He is now a graduate in School of Integrated Circuits, University of Chinese Academy of Sciences. His current research interests include deep learning, neuromorphic computation, and data mining. 
\\
\\
\endbio
\bio{author4}
\textbf{Shushan Qiao} (Member, IEEE) received the B.S. degree in electronics from Hunan University, Changsha, China, in 2003, and the Ph.D. degree in the Institute of Microelectronics of Chinese Academy of Sciences, Beijing, China, in 2008.
He has been a Full Professor since 2018 with the Institute of Microelectronics of Chinese Academy of Sciences. His research interests are in compute-in-memory, artificial intelligence, ultra-low-power processors, and intelligent Microsystems.
\endbio
\bio{author5}
\textbf{Yumei Zhou} received the B.S. degree in electronics from Tsinghua University, Beijing, China, in 1985. Since 1997, she has been a Full Professor with the Institute of Microelectronics of Chinese Academy of Sciences, Beijing. Her research interests include integrated circuit design technology, reliability technology, low-power circuit design, and high-speed interface circuits.
\\
\endbio

\end{document}